# Eeny, meeny, miny, moe. How to choose data for morphological inflection

**Saliha Muradoğlu**[α κ] **Mans Hulden**[χ]
[α]The Australian National University (ANU)   [χ]University of Colorado
[κ]ARC Centre of Excellence for the Dynamics of Language (CoEDL)
saliha.muradoglu@anu.edu.au, mans.hulden@colorado.edu

## Abstract

Data scarcity is a widespread problem in numerous natural language processing (NLP) tasks for low-resource languages. Within morphology, the labour-intensive work of tagging/glossing data is a serious bottleneck for both NLP and language documentation. Active learning (AL) aims to reduce the cost of data annotation by selecting data that is most informative for improving the model. In this paper, we explore four sampling strategies for the task of morphological inflection using a Transformer model: a pair of oracle experiments where data is chosen based on whether the model already can or cannot inflect the test forms correctly, as well as strategies based on high/low model confidence, entropy, as well as random selection. We investigate the robustness of each strategy across 30 typologically diverse languages. We also perform a more in-depth case study of Natügu. Our results show a clear benefit to selecting data based on model confidence and entropy. Unsurprisingly, the oracle experiment, where only incorrectly handled forms are chosen for further training, which is presented as a proxy for linguist/language consultant feedback, shows the most improvement. This is followed closely by choosing low-confidence and high-entropy predictions. We also show that despite the conventional wisdom of larger data sets yielding better accuracy, introducing more instances of high-confidence or low-entropy forms, or forms that the model can already inflect correctly, can reduce model performance.

## 1 Introduction

The need for linguistically annotated data sets is a drive that unites many fields within linguistics. Computational linguists often use labelled data sets for developing NLP systems. Theoretical linguists may utilise corpora for constructing statistical argumentation to support hypotheses about language or phenomena. Documentary linguists create interlinear glossed texts (IGTs) to preserve linguistic and cultural examples, which typically aids in generating a grammatical description. With the renewed focus on low-resource languages and diversity in NLP and the urgency propelled by language extinction, there is widespread interest in addressing this bottleneck.

One method for reducing annotation costs is active learning (AL). AL is an iterative process to optimise model performance by choosing the most critical examples to label. It has been successfully employed for various applications through NLP tasks including deep pre-trained models (BERT) (Ein-Dor et al., 2020), semantic role labelling (Myers and Palmer, 2021), named entity recognition (Shen et al., 2017), word sense disambiguation (Zhu and Hovy, 2007), sentiment classification (Dong et al., 2018) and machine translation (Zeng et al., 2019; Zhang et al., 2018). The iterative nature of AL aligns nicely with the language documentation process. It can be tied into the workflow of a field linguist who consults with a language informant or visits a field site in a periodic manner. Prior to a field trip, a linguist typically prepares material/questions (such as elicitation's or picture tasks[1]) for language consultants which may focus on elements of the language they are working to describe or for material creation (e.g., pedagogical). We propose AL as a method which can provide a supplementary line of insight into the data collection process, particularly for communities that wish to develop and engage with language technology and/or resource building.

Previous work by Palmer (2009) details the efficiency gains from AL in the context of language documentation for the task of morpheme labelling. With deep learning models leading performance for the task of morphological analysis (Pimentel et al., 2021; Vylomova et al., 2020; McCarthy et al.,

---

[1]Or indeed any materials such as those complied by the Max Planck Institute for Psycholinguistics at http://fieldmanuals.mpi.nl/

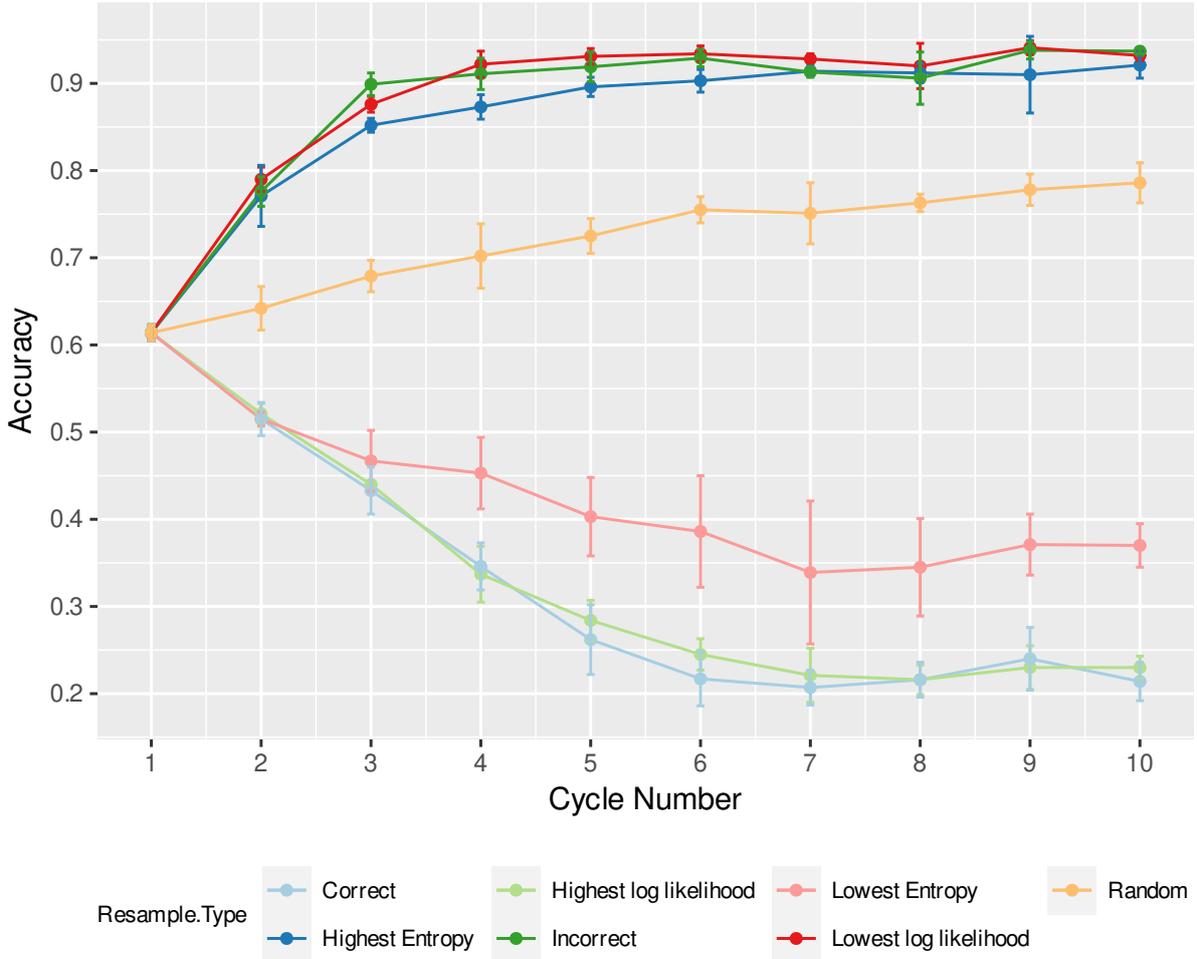

Figure 1: The accuracy for each trained modelled, starting from the baseline (cycle 1). Each cycle 250 instances are re-sampled via the seven sampling methods: correct/incorrect, high/low model confidence, high/low entropy and random (coded with colour). The reported error bars are calculated across 3 separate runs. See Table 1 in Appendix for more detail. After cycle 2, the same sampling strategy is applied to that stream of experiment - e.g. for the lowest log-likelihood strategy, from cycle 2 to 10 the same strategy is used.

2019), AL in the context of neural methods is needed.

This paper addresses the following question: How can we identify the type of data needed to improve model performance? To answer this, we explore the use of AL for the task of morphological inflection using a Transformer model. We run AL simulation experiments with four different sampling strategies: (1) correctness oracle, (2) model confidence, (3) entropy and (4) random selection. These strategies are tested across 30 typologically diverse languages and a 10-cycle iterative experiment using Natügu as a case study.

## 2 Data

We use data from the UniMorph Project (McCarthy et al., 2020), Interlinear Glossed Texts (IGT) from Moeller et al. (2020) and SIGMORPHON (Vylomova et al., 2020; Pimentel et al., 2021). In addition to the data availability, we consider typological diversity when selecting languages to include. Broadly, we attempt to include types of languages that exhibit varying degrees of complexity for inflection. We also consider morphological characteristics coded in WALS; prefixing vs. suffixing (Dryer, 2013), inflectional synthesis of the verb (Bickel and Nichols, 2013b) and exponence (Bickel and Nichols, 2013a). An additional consideration is the paradigm size for the morphological system modelled.

We note data source type to account for the variation in standard across Wikipedia, IGT field data, glossed examples from grammars and data generated from computational grammars.

## 3 Experimental Setup

We train the model as if we were addressing an 'inflection' task (Vylomova et al., 2020). The data is in the form of triplets: lexeme, morphosyntactic tags and the desired output inflected form (e.g. ⟨walk, V;PST, walked⟩)[2]. Each model is trained with the fairseq Transformer (Ott et al., 2019) and our hyperparameters follow Liu and Hulden (2020).

A baseline model is trained, after which more examples are resampled from the baseline test file using the methods detailed below. The initial baseline model is trained with 3,500 instances, 1,000 test and 500 for development. We resample 250 instances.

### 3.1 Sampling strategies

**Oracle** The oracle experiments serve as a proxy for linguist/language expert feedback. 250 examples are sampled based on whether the predicted form is correct/incorrect. The initial filter is supplemented with the following criteria: (1) if there are fewer than 250 incorrect forms, the remaining slots are filled in accordance with examples that exhibit the smallest difference between the first and second output form's log-likelihood, (2) in the case of more than 250 incorrect forms, the incorrect instances are ranked based on the maximum Levenshtein distance between the predicted and target forms. The same selection criteria are applicable for the counterpart correct experiment, with reversed limits (e.g. in the case of less than 250 correct forms, the instances with the largest difference between the first and second log-likelihood are considered).

**Model Confidence** The instances introduced to the training data are sampled based on the model confidence for each form. In this particular strategy, we only record the log-likelihood for the highest-ranked prediction in the beam.

We further examine the correlation between the log-likelihood (continuous variable) and accuracy (dichotomous variable) of the best prediction generated by the model by calculating the Point-Biserial Correlation Coefficient (PBCC). Across the 30 languages we study, the average PBCC is 0.388. Like all correlation coefficients, the PBCC measures the strength of the correlation, and the reported value ranges from -1 to +1, where -1 indicates an inverse association, +1 indicates a positive association, and 0 indicates no association at all.

[2]Data and code available at https://github.com/smuradoglu/ALmorphinfl

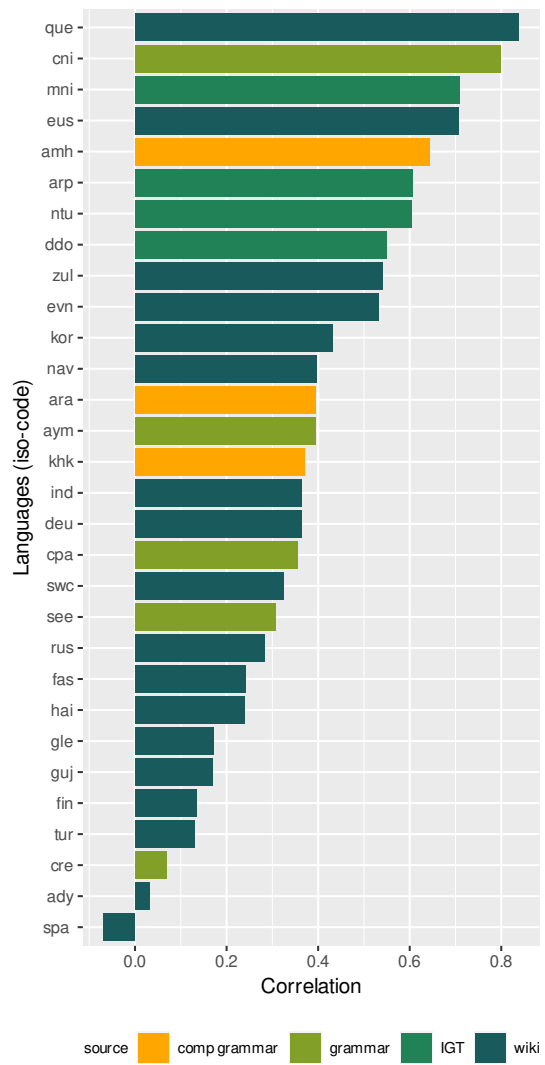

Figure 2: The calculated Point-Biserial Correlation Coefficient (PBCC) between correct prediction and the model log-likelihood, across 30 different languages. The source of the data is also noted with colour.

**Entropy** Here we expand upon the previous strategy—model confidence. We consider the distribution of the ranked output predictions for a particular input and approximate its entropy $-\sum_i p_i \log(p_i)$, by only considering such predictions where $p_i \geq 0.05$, i.e. we calculate $-\sum p_i \log(p_i)$, for all $p_i \geq 0.05$. The model generated log-likelihoods are converted to probabilities and renormalised across the outputs generated by beam search. $p_i = \frac{p_n}{\sum_{j=1}^{b} p_n}$, $b$ being the number of predictions we retrieve from the beam search.

**Random** We contrast the previous methods for re-sampling with random data selection. To establish whether the change in accuracy is statistically significant, we report the average across three independent runs and the standard deviation across the measured accuracy.

## 4 Results and Discussion

To simulate a documentation process, we have chosen Natügu as a case study. The inflection data is from Moeller et al. (2020) and is derived from IGTs—a form that is commonly utilised by field linguists. Our choice of language is further motivated by the morphological complexity exhibited by Natügu. By all accounts Natügu showcases complex morphology (Wurm, 1976; Åshild Næss and Boerger, 2008), particularly on the verb. Historically, this observed complexity led to the language family named as Papuan instead of Austronesian.

Additionally, we observe a positive correlation between prediction correctness and model confidence (0.605). In fact, 4 out of the top 8 correlations (as shown in Figure 2) are languages with IGTs as a data source. For these reasons, we have chosen to examine iterative sampling over 10 cycles.

Figure 1 summarises our results for Natügu. The re-sampling process is iterated over 10 cycles. The first cycle is the baseline/seed run and consists of a 600 instance training set. To account for the impact of random factors affecting the initial training data selection, we have conducted 3 independent seed runs—differing solely on the initial training set. The average accuracy and corresponding standard deviation is reported with the error bars.[3]

The small starting size is motivated by the parallels with language documentation efforts, which are typically a low-resource setting. In each cycle, 250 forms are sampled via the corresponding sampling strategy. By the last cycle the training data consists of 2,850 instances.

Aside from the 3rd and 10th cycle, the lowest log-likelihood sampling consistently provides the greatest improvement. For these two cycles sampling based on incorrect forms outperforms selection based on low confidence. In general, the top 3 selection methods are ranked as follows: low log-likelihood, incorrect and highest entropy forms. We note the possible interplay between paradigm size (907 unique tag combinations) and training size set (1,100 by cycle 3); unseen morphosyntactic categories will be most informative and presumably beneficial to model performance.

Given the strong correlation between prediction accuracy and model confidence for Natügu, we expect similarity in trajectory across cycle number and accuracy for the oracle and model-confidence based sampling strategies. Figure 1 verifies these forecasts; we see that the sampling based on prediction correctness (in light blue) and the sampling based on the highest log-likelihood (in light green) almost look identical. The same is observable for low log-likelihood (in red) and sampling based on incorrect prediction (green).

The lowest log-likelihood sampling method can be seen as an approximation for the highest entropy selection method, and by extension, the highest log-likelihood as an approximation for the lowest entropy selection. Our results for iterative AL for Natügu show that choosing by approximation is a higher risk endevaour. The choice either works really well or not at all. When we contrast low entropy and high model confidence as a selection strategy we can see that low entropy limits the impact of high model confidence since it accounts for a distribution rather than the single value approximation. We observe similar behaviour between the the high entropy and low confidence selection strategies. Random sampling shows gradual improvement.

Work by Yuan et al. (2020) highlight the issues with uncertainty sampling for deep learning models; noting that neural networks are poorly calibrated (Guo et al., 2017), and that the correlation between high confidence and correctness is not well established. We explore this correlation for our models in Figure 2. We observe a similar uncertainty with an overall slight positive correlation across the 30 languages examined. Despite this, our results show that data selection based on low model confidence yields significant improvement of model accuracy. The work presented here is intended as a preliminary baseline; we leave it to future work to consider calibration methods such as temperature scaling.

Interestingly, despite an increase in training data size, introducing new data that the model already can inflect correctly, or low-entropy or high-confidence forms actually reduces model performance despite the widely-held notion that more data is better. Another recent study by Samir and

---
[3]Individual values can be found in Table 1 of the Appendix.

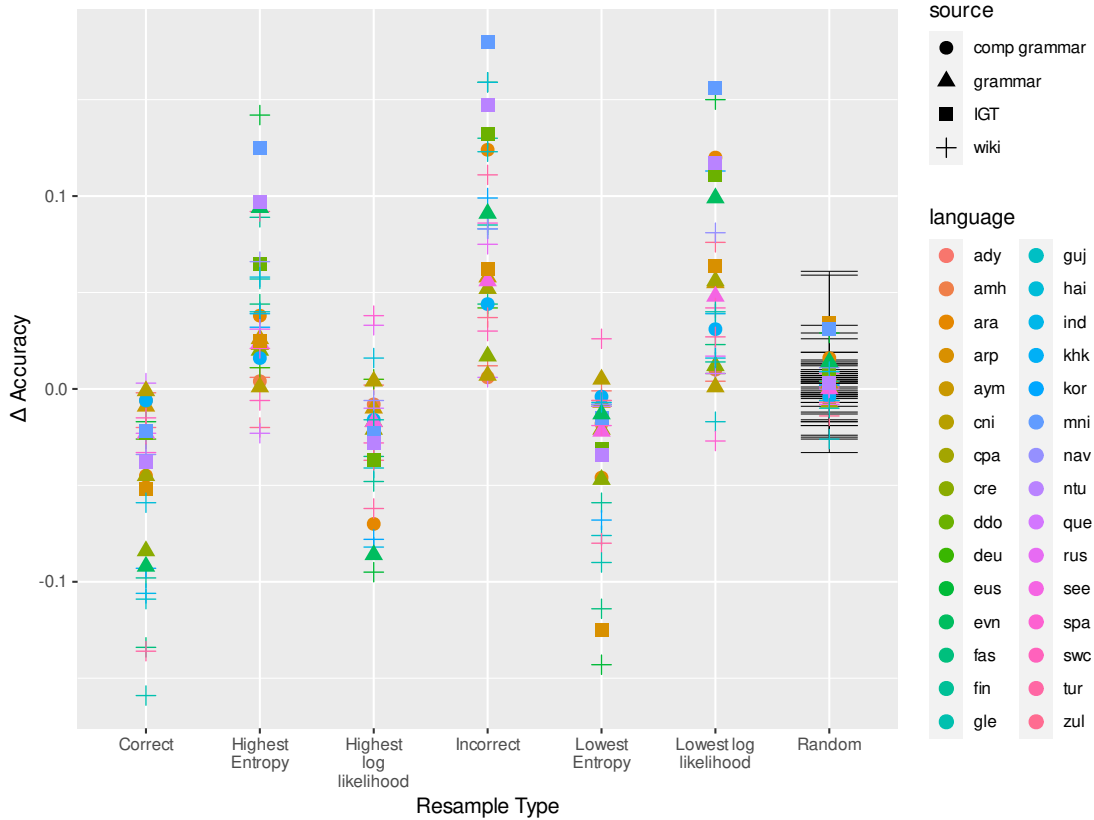

Figure 3: The change in accuracy (from the established baseline) is reported with each sampling strategy, across 30 different languages (coded with colour). The source of data is also noted with tick shapes.

Silfverberg (2022) reports similar behaviour, where data hallucination reduces prediction accuracy for words that exhibited reduplication.

We extend the same sampling strategies to 30 different languages for one round of re-training. The results are summarised in Figure 3. Within the 30 languages we ensure to include languages with large inflection table sizes (ranging from 12 to 700+), different scripts (Latin, Cyrillic, Arabic, Hangul, Ge'ez and Gujarati) and morphological typology (agglutinating, fusional, polysynthetic). We code for the source of the data, and see no particular deviation from the overall observed behaviour. The reported error bars for random sampling correspond to the standard deviation across three independent runs of random sampling.

It is clear that in general, the sampling strategies can be ordered for prediction accuracy improvement in the following manner: incorrect, lowest log-likelihood, highest entropy, random, highest log-likelihood, lowest entropy and finally correct form sampling. While a handful of languages deviate from this pattern (e.g. Swahili or Dido),[4] it holds true for a majority of the languages considered.

## 5 Conclusion

In this paper we examine four different sampling strategies within an AL framework for modelling morphological inflection using a Transformer model. We consider correct/incorrect prediction, model confidence, entropy and random selection as sampling strategies. Our results clearly show that AL can significantly improve learning rates for morphological inflection. Unsurprisingly, adding oracle-indicated incorrect forms for training yields the greatest model improvement. In the absence of a language expert, model confidence can be used to prioritise data annotation. This holds true across 30 different languages. We also show that larger datasets do not always yield better results; the diversity of the training set matters.

Future research should extend the analysis to incorporate language-specific factors—such as model performance for each morphosyntactic slot within the morphological paradigm.

---

[4] see Table.3 in Appendix for more detail.

## Limitations

The primary limitation of this study is that the results are not evaluated in a real life documentation scenario. While we have tried to address this gap by noting the source of data, and have enlisted IGT data to serve as a proxy, we acknowledge that fieldwork data is often inconsistent, noisy and requires much more data cleaning. The data used for these experiments is, for the most part, already structured as a paradigm.

In addition, the simple metric of accuracy can be crude and is often prone to some degree of fluctuation. To minimise these effects we have considered the change in accuracy across sampling cycles instead. Lastly, we have tried to collate a diverse set of languages to consider. However, this is largely limited by the availability of data. It is likely that several morphophonological phenomena are not included within the data sets used here.

## A  Appendix

|  |  | **Resample by:** | | | | | | | | | | | | | | |
|---|---|---|---|---|---|---|---|---|---|---|---|---|---|---|---|---|
|  |  | Lowest $log(p_i)$ | | | | | Highest $log(p_i)$ | | | | | Random | | | | |
| Cycle # | training size | S1 | S2 | S3 | avg | std | S1 | S2 | S3 | avg | std | S1 | S2 | S3 | avg | std |
| 1 | 600 | 0.618 | 0.621 | 0.604 | 0.614 | 0.009 | 0.618 | 0.621 | 0.604 | 0.614 | 0.009 | 0.618 | 0.621 | 0.604 | 0.614 | 0.009 |
| 2 | 850 | 0.800 | 0.795 | 0.774 | 0.790 | 0.014 | 0.508 | 0.532 | 0.522 | 0.521 | 0.012 | 0.617 | 0.666 | 0.642 | 0.642 | 0.025 |
| 3 | 1100 | 0.870 | 0.872 | 0.886 | 0.876 | 0.009 | 0.439 | 0.442 | 0.439 | 0.440 | 0.002 | 0.679 | 0.697 | 0.662 | 0.679 | 0.018 |
| 4 | 1350 | 0.927 | 0.933 | 0.905 | 0.922 | 0.015 | 0.374 | 0.317 | 0.321 | 0.337 | 0.032 | 0.723 | 0.723 | 0.659 | 0.702 | 0.037 |
| 5 | 1600 | 0.922 | 0.939 | 0.932 | 0.931 | 0.009 | 0.275 | 0.310 | 0.267 | 0.284 | 0.023 | 0.727 | 0.744 | 0.705 | 0.725 | 0.020 |
| 6 | 1850 | 0.934 | 0.925 | 0.943 | 0.934 | 0.009 | 0.266 | 0.234 | 0.236 | 0.245 | 0.018 | 0.741 | 0.771 | 0.754 | 0.755 | 0.015 |
| 7 | 2100 | 0.921 | 0.929 | 0.933 | 0.928 | 0.006 | 0.256 | 0.198 | 0.208 | 0.221 | 0.031 | 0.735 | 0.726 | 0.791 | 0.751 | 0.035 |
| 8 | 2350 | 0.940 | 0.890 | 0.929 | 0.920 | 0.026 | 0.236 | 0.207 | 0.206 | 0.216 | 0.017 | 0.756 | 0.774 | 0.758 | 0.763 | 0.010 |
| 9 | 2600 | 0.943 | 0.932 | 0.948 | 0.941 | 0.008 | 0.225 | 0.208 | 0.258 | 0.230 | 0.025 | 0.758 | 0.783 | 0.794 | 0.778 | 0.018 |
| 10 | 2850 | 0.929 | 0.927 | 0.939 | 0.932 | 0.006 | 0.242 | 0.230 | 0.217 | 0.230 | 0.013 | 0.760 | 0.795 | 0.803 | 0.786 | 0.023 |

Table.1: Model accuracies for iterative sampling for Natügu, across the lowest and highest low-likelihoods and random sampling strategies. S1, S2, S3 corresponds to seed 1, seed 2 and seed 3 respectively. Avg and std indicate the average value across the three seed runs and the standard deviation. Data used to generate Figure.1.

|  |  | **Resample by:** | | | | | | | | | | | | | | | | | | | |
|---|---|---|---|---|---|---|---|---|---|---|---|---|---|---|---|---|---|---|---|---|---|
|  |  | Incorrect | | | | | Correct | | | | | Highest Entropy | | | | | Lowest Entropy | | | | |
| Cycle # | training size | S1 | S2 | S3 | avg | std | S1 | S2 | S3 | avg | std | S1 | S2 | S3 | avg | std | S1 | S2 | S3 | avg | std |
| 1 | 600 | 0.618 | 0.621 | 0.604 | 0.614 | 0.009 | 0.618 | 0.621 | 0.604 | 0.614 | 0.009 | 0.618 | 0.621 | 0.604 | 0.614 | 0.009 | 0.618 | 0.621 | 0.604 | 0.614 | 0.009 |
| 2 | 850 | 0.778 | 0.791 | 0.758 | 0.776 | 0.017 | 0.530 | 0.521 | 0.493 | 0.515 | 0.019 | 0.792 | 0.790 | 0.731 | 0.771 | 0.035 | 0.506 | 0.520 | 0.518 | 0.515 | 0.008 |
| 3 | 1100 | 0.896 | 0.887 | 0.913 | 0.899 | 0.013 | 0.452 | 0.446 | 0.402 | 0.433 | 0.027 | 0.848 | 0.861 | 0.848 | 0.852 | 0.008 | 0.507 | 0.445 | 0.448 | 0.467 | 0.035 |
| 4 | 1350 | 0.901 | 0.931 | 0.900 | 0.911 | 0.018 | 0.373 | 0.346 | 0.320 | 0.346 | 0.027 | 0.876 | 0.857 | 0.885 | 0.873 | 0.014 | 0.500 | 0.439 | 0.421 | 0.453 | 0.041 |
| 5 | 1600 | 0.923 | 0.934 | 0.901 | 0.919 | 0.017 | 0.291 | 0.278 | 0.216 | 0.262 | 0.040 | 0.892 | 0.908 | 0.888 | 0.896 | 0.011 | 0.454 | 0.386 | 0.369 | 0.403 | 0.045 |
| 6 | 1850 | 0.935 | 0.935 | 0.917 | 0.929 | 0.010 | 0.212 | 0.189 | 0.250 | 0.217 | 0.031 | 0.895 | 0.895 | 0.918 | 0.903 | 0.013 | 0.460 | 0.342 | 0.357 | 0.386 | 0.064 |
| 7 | 2100 | 0.906 | 0.916 | 0.916 | 0.913 | 0.006 | 0.189 | 0.229 | 0.203 | 0.207 | 0.020 | 0.919 | 0.916 | 0.908 | 0.914 | 0.006 | 0.362 | 0.248 | 0.408 | 0.339 | 0.082 |
| 8 | 2350 | 0.929 | 0.872 | 0.917 | 0.906 | 0.030 | 0.234 | 0.220 | 0.194 | 0.216 | 0.020 | 0.919 | 0.917 | 0.899 | 0.912 | 0.011 | 0.408 | 0.301 | 0.327 | 0.345 | 0.056 |
| 9 | 2600 | 0.928 | 0.947 | 0.940 | 0.938 | 0.010 | 0.210 | 0.229 | 0.280 | 0.240 | 0.036 | 0.933 | 0.860 | 0.938 | 0.910 | 0.044 | 0.365 | 0.409 | 0.339 | 0.371 | 0.035 |
| 10 | 2850 | 0.935 | 0.940 | 0.937 | 0.937 | 0.003 | 0.229 | 0.188 | 0.224 | 0.214 | 0.022 | 0.933 | 0.904 | 0.925 | 0.921 | 0.015 | 0.378 | 0.389 | 0.342 | 0.370 | 0.025 |

Table.2: Model accuracies for iterative sampling for Natügu, across incorrect, correct, highest and lowest entropy sampling strategies. S1, S2, S3 corresponds to seed 1, seed 2 and seed 3 respectively. Avg and std indicate the average value across the three seed runs and the standard deviation. Data used to generate Figure.1.

| Language | Iso-code | PBCC | p-value |
|---|---|---|---|
| Adyghe | ady | 0.031 | 3.29E-01 |
| Amharic | amh | 0.643 | 5.74E-118 |
| Arabic | ara | 0.394 | 1.53E-38 |
| Arapaho | arp | 0.607 | 1.08E-101 |
| Aymara | aym | 0.394 | 1.64E-38 |
| Asháninka | cni | 0.799 | 1.10E-222 |
| Palantla Chinantec | cpa | 0.355 | 4.10E-31 |
| Cree | cre | 0.069 | 2.91E-02 |
| Dido | ddo | 0.550 | 4.24E-80 |
| German | deu | 0.363 | 1.60E-32 |
| Basque | eus | 0.707 | 3.67E-152 |
| Evenki | evn | 0.532 | 2.94E-74 |
| Persian | fas | 0.242 | 7.68E-15 |
| Finnish | fin | 0.135 | 1.84E-05 |
| Irish | gle | 0.172 | 4.24E-08 |
| Gujarati | guj | 0.168 | 8.40E-08 |
| Haida | hai | 0.240 | 1.47E-14 |
| Indonesian | ind | 0.364 | 1.05E-32 |
| Halh Mongolian | khk | 0.370 | 8.44E-34 |
| Korean | kor | 0.431 | 1.78E-46 |
| Manipuri | mni | 0.709 | 2.34E-153 |
| Navaho | nav | 0.396 | 7.66E-39 |
| Natügu | ntu | 0.605 | 1.13E-100 |
| Quechua | que | 0.838 | 3.44E-265 |
| Russian | rus | 0.282 | 9.57E-20 |
| Seneca | see | 0.306 | 3.75E-23 |
| Spanish | spa | -0.069 | 3.02E-02 |
| Swahili | swc | 0.324 | 6.16E-26 |
| Turkish | tur | 0.129 | 4.17E-05 |
| Zulu | zul | 0.540 | 9.23E-77 |

Table.2: Correct and model log-likelihood correlation based on baseline for each language. The reported value is a Point-Biserial Correlation Coefficient (PBCC) with the respective p-value. Data used to generate Figure.2.

| Language | Source | # Tables | Baseline | Resample by: | | | | | | $Random_1$ | $Random_2$ | $Random_3$ | $Random_{avg}$ | ± Std dev |
| | | | | Correct | Incorrect | Lowest $log(p_i)$ | Highest $log(p_i)$ | Lowest Entropy | Highest Entropy | | | | | |
| --- | --- | --- | --- | --- | --- | --- | --- | --- | --- | --- | --- | --- | --- | --- |
| ady | Wiki | 430 | 0.986 | 0.984 | 0.998 | 0.990 | 0.988 | 0.985 | 0.992 | 0.989 | 0.990 | 0.990 | 0.990 | 0.001 |
| amh | c. grammar | 285 | 0.983 | 0.977 | 0.989 | 0.993 | 0.975 | 0.965 | 0.987 | 0.980 | 0.977 | 0.977 | 0.978 | 0.002 |
| ara | c. grammar | 83 | 0.800 | 0.755 | 0.924 | 0.920 | 0.730 | 0.754 | 0.838 | 0.818 | 0.805 | 0.824 | 0.816 | 0.010 |
| aym | grammar | 55 | 0.933 | 0.924 | 0.991 | 0.988 | 0.923 | 0.926 | 0.959 | 0.939 | 0.941 | 0.938 | 0.939 | 0.002 |
| cpa | grammar | 490 | 0.843 | 0.798 | 0.895 | 0.899 | 0.822 | 0.822 | 0.866 | 0.820 | 0.857 | 0.832 | 0.836 | 0.019 |
| cre | grammar | 22 | 0.113 | 0.029 | 0.130 | 0.125 | 0.096 | 0.066 | 0.133 | 0.110 | 0.116 | 0.115 | 0.114 | 0.003 |
| deu | wiki | 450 | 0.937 | 0.911 | 0.979 | 0.945 | 0.942 | 0.928 | 0.948 | 0.935 | 0.939 | 0.920 | 0.931 | 0.010 |
| eus | wiki | 12 | 0.755 | 0.738 | 0.914 | 0.905 | 0.660 | 0.612 | 0.897 | 0.813 | 0.754 | 0.784 | 0.784 | 0.030 |
| fas | wiki | 39 | 0.178 | 0.044 | 0.222 | 0.201 | 0.143 | 0.064 | 0.222 | 0.186 | 0.182 | 0.178 | 0.182 | 0.004 |
| fin | wiki | 97 | 0.587 | 0.489 | 0.717 | 0.601 | 0.539 | 0.528 | 0.676 | 0.593 | 0.584 | 0.590 | 0.589 | 0.005 |
| guj | wiki | 280 | 0.620 | 0.511 | 0.743 | 0.603 | 0.579 | 0.544 | 0.660 | 0.601 | 0.587 | 0.594 | 0.594 | 0.007 |
| ind | wiki | 750 | 0.551 | 0.445 | 0.634 | 0.590 | 0.469 | 0.543 | 0.590 | 0.556 | 0.530 | 0.549 | 0.545 | 0.013 |
| khk | c. grammar | 720 | 0.936 | 0.930 | 0.980 | 0.967 | 0.920 | 0.932 | 0.952 | 0.944 | 0.934 | 0.918 | 0.932 | 0.013 |
| kor | wiki | 60 | 0.597 | 0.504 | 0.696 | 0.710 | 0.519 | 0.529 | 0.629 | 0.595 | 0.597 | 0.606 | 0.599 | 0.006 |
| rus | wiki | 320 | 0.884 | 0.861 | 0.959 | 0.901 | 0.917 | 0.878 | 0.915 | 0.857 | 0.889 | 0.881 | 0.876 | 0.017 |
| see | grammar | 135 | 0.895 | 0.872 | 0.951 | 0.943 | 0.878 | 0.873 | 0.919 | 0.902 | 0.884 | 0.898 | 0.895 | 0.009 |
| spa | wiki | 75 | 0.880 | 0.847 | 0.966 | 0.853 | 0.918 | 0.861 | 0.901 | 0.884 | 0.874 | 0.884 | 0.881 | 0.006 |
| swc | wiki | 53 | 0.931 | 0.916 | 0.961 | 0.973 | 0.903 | 0.957 | 0.925 | 0.939 | 0.941 | 0.937 | 0.939 | 0.002 |
| tur | wiki | 35 | 0.464 | 0.328 | 0.575 | 0.491 | 0.402 | 0.384 | 0.556 | 0.456 | 0.462 | 0.452 | 0.457 | 0.005 |
| zul | wiki | 62 | 0.881 | 0.861 | 0.918 | 0.957 | 0.844 | 0.875 | 0.861 | 0.876 | 0.868 | 0.856 | 0.867 | 0.010 |
| arp | IGT | 470 | 0.290 | 0.238 | 0.352 | 0.354 | 0.265 | 0.165 | 0.315 | 0.326 | 0.296 | 0.349 | 0.324 | 0.027 |
| que | wiki | 25 | 0.982 | 0.985 | 0.988 | 0.990 | 0.972 | 0.973 | 0.959 | 0.969 | 0.994 | 0.982 | 0.982 | 0.013 |
| gle | wiki | 350 | 0.387 | 0.228 | 0.472 | 0.427 | 0.371 | 0.297 | 0.444 | 0.372 | 0.375 | 0.385 | 0.377 | 0.007 |
| ddo | IGT | 400 | 0.793 | 0.770 | 0.925 | 0.904 | 0.756 | 0.762 | 0.858 | 0.804 | 0.799 | 0.806 | 0.803 | 0.004 |
| nav | wiki | 280 | 0.860 | 0.826 | 0.943 | 0.941 | 0.854 | 0.852 | 0.926 | 0.874 | 0.862 | 0.877 | 0.871 | 0.008 |
| mni | IGT | 525 | 0.752 | 0.730 | 0.932 | 0.908 | 0.729 | 0.737 | 0.877 | 0.784 | 0.784 | 0.780 | 0.783 | 0.002 |
| evn | grammar | 2250 | 0.460 | 0.368 | 0.551 | 0.559 | 0.374 | 0.447 | 0.554 | 0.470 | 0.473 | 0.479 | 0.474 | 0.005 |
| cni | grammar | 105 | 0.992 | 0.991 | 0.999 | 0.993 | 0.996 | 0.997 | 0.993 | 0.993 | 0.996 | 0.995 | 0.995 | 0.002 |
| hai | wiki | 31 | 0.715 | 0.656 | 0.874 | 0.731 | 0.731 | 0.708 | 0.773 | 0.728 | 0.727 | 0.717 | 0.724 | 0.006 |
| ntu | IGT | 560 | 0.800 | 0.762 | 0.947 | 0.917 | 0.772 | 0.766 | 0.897 | 0.811 | 0.792 | 0.806 | 0.803 | 0.010 |

Table.3: Model accuracies for each sampling strategy, across 30 different languages. Data used to generate Figure.3.